# Corpus Similarity Measures Remain Robust Across Diverse Languages

Draft

Haipeng Li and Jonathan Dunn

This paper experiments with frequency-based corpus similarity measures across 39 languages using a register prediction task. The goal is to quantify (i) the distance between different corpora from the same language and (ii) the homogeneity of individual corpora. Both of these goals are essential for measuring how well corpus-based linguistic analysis generalizes from one dataset to another. The problem is that previous work has focused on Indo-European languages, raising the question of whether these measures are able to provide robust generalizations across diverse languages. This paper uses a register prediction task to evaluate competing measures across 39 languages: how well are they able to distinguish between corpora representing different contexts of production? Each experiment compares three corpora from a single language, with the same three digital registers shared across all languages: social media, web pages, and Wikipedia. Results show that measures of corpus similarity retain their validity across different language families, writing systems, and types of morphology. Further, the measures remain robust when evaluated on out-of-domain corpora, when applied to low-resource languages, and when applied to different sets of registers. These findings are significant given our need to make generalizations across the rapidly increasing number of corpora available for analysis.

*Word Count*: 8,685

# 1. Corpus Similarity in a Cross-Linguistic Setting

The basic problem of corpus similarity is to find the distance between two corpora. This conception of distance is not confined to one part of the linguistic signal: differences between corpora could include lexical, morphological, syntactic, or even semantic properties. The goal of a corpus similarity measure is to provide a broad representation of distance which aggregates across specific linguistic features (Kilgarriff, 2001). The challenge is that this broad conception of linguistic similarity could be subject to cross-linguistic variation. For example, sequence-based measures might capture syntactic differences in an analytic language but capture morphological differences in an agglutinative language. The question here is whether corpus similarity measures remain robust across languages from different families with different types of morphology and different writing systems. This question is important because the number of corpora available for linguistic analysis has been growing rapidly: depending on the language in question, we now have hundreds to thousands of corpora available. Any attempt to generalize a corpus-based analysis beyond one specific corpus depends on corpus similarity measures; we are otherwise left with a large number of isolated corpus-specific findings. The experiments in this paper establish the robustness of the core methods for measuring how well corpus-based linguistic analysis generalizes from one dataset to another.

The main gap in our understanding of corpus similarity comes from the fact that most work is focused entirely on English. How much confidence can we have when applying these measures to new languages with very different linguistic properties? How much confidence can we have when applying measures to new domains or to low-resource languages that have limited training data? This is an important question because it could be the case that frequency-based similarity measures depend on superficial properties, so that they would not generalize well across languages. The experiments in this paper show that there are limited variations in accuracy across a set of 39 languages, once the best parameters for each language have been discovered. The experiments further show that the measures are generally robust to out-of-domain feature selection and that accuracy remains high when evaluated on different sets of corpora. This means that we can have confidence in these measures when they are applied to new contexts and new corpora. These are important findings which show that corpus similarity measures are not one-off *ad hoc* representations that depend on specific languages, specific corpora, or specific contexts of production.

This paper makes four contributions: *First*, we create comparable corpora across 39 languages that represent three distinct registers (social media, web pages, and Wikipedia). We describe the selection and creation of this corpus in Section 3 and ask whether each of these data sources represents a single unique register in Section 5. *Second*, we propose several pairwise similarity measures which take as input two subsets of a corpus, each subset containing at least 10k words. These measures include and expand upon previous work on frequency-based corpus similarity. *Third*, we provide a rigorous evaluation of these measures across all 39 languages to reach robust cross-linguistic generalizations about how well these measures work. *Fourth*, we provide a Python package which implements these measures. The larger linguistic contribution of this paper is to show that corpus similarity measures remain robust across language families, across types of morphology, and across different writing systems. This robustness, in turn, is important because it allows corpus-based linguistic analysis to make generalizations across the increasingly large number of corpora that are available for analysis.

Corpus similarity measures are increasingly important for quantitative corpus-based linguistics because they allow us to make generalizations across corpora. For example, imagine a study of particle placement in news articles drawn from different countries. Such a study would likely find significant syntactic differences across geographic locations (c.f., Dunn, 2019). But before we generalized these findings we would need to answer two important questions: First, would the results have changed if we had observed a different sub-set of news articles? Given the enormous number of publications and articles in English, any corpus-based study works with a very limited sample. Corpus homogeneity measures can be used to determine how much variation there is within the same context (i.e., news articles from the US). Second, how much would a corpus of news articles from different countries tell us about the dialect of English used in those countries? Given the enormous number of registers that represent each dialect (social media, news broadcasts, novels, spoken language, etc), we can use corpus similarity measures to estimate how well findings from one corpus (news articles) generalize to a new corpus (social media). The measures evaluated in this paper are essential for making generalizations in corpus linguistics (c.f., Kilgarriff, 2001). Previous work, by focusing largely on English, leaves open the question of how well we can make corpus-based generalizations from non-English corpora.

### *1.1. Defining Digital Registers*

We focus on register variation as a test case for corpus similarity because register is a significant source of linguistic variation in corpora (Biber, 2012). This paper develops an accuracy metric that is based on predicting whether two samples are drawn from the same register or from different registers. Because each language is represented using comparable corpora from the same three registers, we are able to compare the experimental results across languages. The term *register* here refers to the context of production: for example, the author, the audience, and the communicative purpose behind a corpus. Register is related to *genre*, although genre usually refers to more formal properties of a document, such as an opening greeting in an email or a location tag in a news article (Biber & Conrad, 2012). We focus on register as a test case because register itself is the most common dimension of difference between corpora (i.e., news articles vs social media).

Each register has its own situational properties. For example, the author of a tweet is a single known individual while the author of a web page is an unknown individual and a Wikipedia article is drawn from many contributors. The communicative purpose of Wikipedia is to present information in an objective manner; social media, however, has a number of distinct purposes: communicating with friends, making announcements, etc. The main experiments in this paper rely on three digital registers: social media, web pages, and Wikipedia articles. Each of these registers has its own situational properties. We evaluate, in Section 5.4, whether the results of our experiments depend on a this specific mix of registers; the results show that the same generalizations remain across a distinct set of six non-digital registers.

## 2. Measures of Corpus Similarity

This section reviews previous work on corpus similarity and motivates the experimental framework taken in this paper. The basic problem was originally outlined by Kilgarriff (2001), whose study showed that a frequency-based approach performs best. This type of evaluation was extended in later work (Fothergill et al., 2016), which explored much larger corpora. These larger corpora allow their experiments to also consider the effect of corpus size on similarity measures. Three measures are compared: $\chi^2$ similarity, the perplexity of trained language models, and topic similarity from trained topic models. Overall, the model-based methods do not perform as well as the frequency-based $\chi^2$ measure, replicating Kilgarriff's original finding. The expanded experiments involving corpus size further show that the best accuracy is observed with 4k features, substantially higher than Kilgarriff's original 500.

More recently, frequency-based measures were used (Dunn, 2021) to show that there is a consistent agreement between digital corpora (the web and tweets) across nine languages and 84 language varieties. In this work, word unigram frequencies and character trigram frequencies are used to calculate Spearman's $\rho$; the resulting similarity values then measure variation within and between registers for different language varieties. Related work has also studied the impact on corpus similarity measurements caused by different choices for corpus size (Piperski, 2018). The results of these experiments, using the British National Corpus, show that Euclidean distance is least influenced by corpus size and thus is best suited for the purpose of comparing corpora when it is necessary to compare samples of different sizes. Other work uses Jensen-Shannon divergence (JSD) to conduct corpus comparison experiments (Lu et al., 2020). While this method is based on word frequency, there are no quantitative values produced for corpus similarity; therefore, we do not consider JSD in this paper. Other recent work proposes the Sum of Minimum Frequencies (SMF) as a new corpus similarity measure (Piperski, 2017). This method is based on word frequencies and outperforms perplexity-based measures, as expected. However, the results are comparable to Spearman's $\rho$ and $\chi^2$. Thus, we do not include it in this study. It is important to note that corpus similarity measures based on the $\chi^2$ are not interpreted as significance tests, so that we do not need to consider limitations on the interpretation of such significance tests (c.f., Mačutek & Wimmer, 2013; Wallis, 2013).

Another related line of work is the detection of similarity between documents or articles, with a focus on topic. For example, in recent work (Nanayakkara & Ranathunga, 2018) cosine similarity has been used to group news articles. Individual similarity values are compared with a threshold value in order to cluster articles into groups of related topics. In other recent work, (Torres-Moreno et al., 2014), corpus similarity is evaluated for the task of paraphrase detection in German. This problem is a more specific form of the deduplication problem, in which the goal is to find parts of a corpus which are very similar. The difference with corpus similarity, however, is that paraphrase detection is focused on relatively short spans of a document while corpus similarity is applied to a large number of documents in the aggregate. Recent work has also examined the relationship between corpus-based measures and readability (Pires, et al., 2017). It remains a open question whether similar corpora have similar readability scores.

In the current paper, we validate similarity measures using their accuracy for predicting whether a pair of samples (two corpora) are from the same register or from different registers. Other recent work has approached this problem of finding thresholds for similarity measures (Ali, 2011, Leban et al., 2016). We take a similar thresholding approach, as described in Section 4.1, converting a continuous similarity measure into a binary label for use in a prediction task.

Given the use of thresholds here, we could also compare corpus similarity to text classification. A major distinction between the two is that text classification is a supervised problem, with at least some amount of training data required. For corpus similarity measures, however, samples are compared using only frequency ranks. This means, for example, that there are no feature weights to optimize using training data. Most text classification tasks are performed within a single register, rather than across registers. Multi-lingual text classification on smaller documents remains a challenge, especially within a single register (c.f., Mutuvi, et al., 2020). At the same time, recent work has shown that register-specific word frequency distributions are helpful for focusing models in an educational domain (Ehara, 2019). This combination of corpus similarity within a larger classification problem shows the importance of being able to navigate the relationships between large numbers of corpora. The range of work reviewed in this section shows the importance of corpus similarity as a means of understanding the relationships between different corpora, a problem that is relevant to corpus linguistics, computational linguistics, and experimental linguistics.

## 3. Data and Methodology

The corpora for this study are drawn from three digital registers: social media (TW: from Twitter), web pages (CC: from the *Corpus of Global Language use*: Dunn, 2020), and Wikipedia (WK: from the Wikipedia dump of March 2020). For TW and CC, data is sorted by language using the *idNet* language identification package (Dunn, 2020). For WK, the language label is derived from the Wikipedia domain. The TW corpora is taken to represent the social media register and the WK corpora is taken to represent the non-fiction or encyclopedic article register. We know, however, that the CC corpora contains a number of potentially distinct sub-registers like forum posts or comments on news articles (Sardinha, 2018). These unlabeled sub-registers add an important dimension to these experiments: are the CC corpora more heterogeneous, as we would expect given that they contain these potential sub-registers? Is the level of heterogeneity consistent across languages, or does the distinctiveness of internet sub-registers vary across languages? This question is examined further in Section 5.2. If these sources do not constitute unique registers, the overall prediction accuracy will be low.

      The list of languages used is shown in Table 1, along with each language's family, script type, and morphological type. These three classifications for each language are included because these are all factors that may influence the performance of corpus similarity measures. Because Indo-European is a well-represented family, it is divided into branches (for example, *IE: Germanic*).

Languages are divided into four types of writing systems. *Alphabetic* scripts use characters to represent individual phonemes. *Abjad* scripts use characters to represent consonants and leave vowels unrepresented. *Abugida* scripts represent consonant-vowel sequences together. Finally, *logographic* or syllabic scripts use characters that represent an entire word, morpheme, or syllable. The type of writing system may have an influence over corpus similarity measures because there is variation in both the unit of representation (i.e., phoneme vs word) as well as the inventory size (i.e., alphabets have fewer units than logographic systems).

A broad morphological categorization for each language is also included in Table 1. There are four categories: *Agglutinative* languages have a range of different morphemes that retain the same form. We might expect character-based features to detect morphemes in agglutinative languages, for example. *Fusional* languages combine multiple functions (like person and number) into a single morpheme. A third type, *analytic* languages, tend to use grammatical words instead of morphemes. We might expect, for example, that word-based features work better for analytic languages. Finally, we use the term *root-and-pattern* to describe Arabic and Hebrew, which do not fit nicely into the previous typology. The goal for this classification is to enable us to see broad patterns in the performance of different corpus similarity measures. Many of these languages have been studied in isolation (e.g., Hindi in Pande & Dhami, 2013), but not in a systematic cross-linguistic fashion.

Taken together, the 39 languages chosen for this study represent a broad sample of languages: from 15 sub-families, with examples representing all kinds of morphology and writing systems. We expect, then, that the experiments in this paper can be reasonably taken to make generalizations about corpus similarity measures across all languages. This is an important contribution because previous work has been largely confined to English. The final column in Table 1 refers to the best feature type for each language; this is described in detail in Section 4.2.

**Table 1. Languages by Family, Script, and Morphology**

| Name | Code | Family | Script | Morphology | Feature Type |
|---|---|---|---|---|---|
| Vietnamese | vie | Austroasiatic | Alphabet | Analytic | Word 2-grams |
| Indonesian | ind | Austronesian | Alphabet | Agglutinative | Word 1-grams |
| Tagalog | tgl | Austronesian | Alphabet | Agglutinative | Word 1-grams |
| Tamil | tam | Dravidian | Abugida | Agglutinative | Char 4-grams |
| Telugu | tel | Dravidian | Abugida | Agglutinative | Char 4-grams |

| Name | Code | Family | Script | Morphology | Feature Type |
| --- | --- | --- | --- | --- | --- |
| Bulgarian | bul | IE:Balto-Slavic | Alphabet | Fusional | Char 4-grams |
| Czech | ces | IE:Balto-Slavic | Alphabet | Fusional | Char 4-grams |
| Latvian | lav | IE:Balto-Slavic | Alphabet | Fusional | Char 4-grams |
| Polish | pol | IE:Balto-Slavic | Alphabet | Fusional | Char 4-grams |
| Russian | rus | IE:Balto-Slavic | Alphabet | Fusional | Char 4-grams |
| Slovenian | slv | IE:Balto-Slavic | Alphabet | Fusional | Char 4-grams |
| Ukrainian | ukr | IE:Balto-Slavic | Alphabet | Fusional | Char 4-grams |
| Danish | dan | IE:Germanic | Alphabet | Analytic | Char 4-grams |
| German | deu | IE:Germanic | Alphabet | Fusional | Char 4-grams |
| English | eng | IE:Germanic | Alphabet | Analytic | Char 4-grams |
| Dutch | nld | IE:Germanic | Alphabet | Analytic | Char 4-grams |
| Norwegian | nor | IE:Germanic | Alphabet | Analytic | Char 4-grams |
| Swedish | swe | IE:Germanic | Alphabet | Analytic | Char 4-grams |
| Greek | ell | IE:Hellenic | Alphabet | Fusional | Char 4-grams |
| Farsi | fas | IE:Indo-Iranian | Abjad | Analytic | Word 1-grams |
| Hindi | hin | IE:Indo-Iranian | Abugida | Fusional | Char 4-grams |
| Urdu | urd | IE:Indo-Iranian | Abjad | Fusional | Char 4-grams |
| Catalan | cat | IE:Romance | Alphabet | Fusional | Char 4-grams |
| French | fra | IE:Romance | Alphabet | Fusional | Word 1-grams |
| Galician | glg | IE:Romance | Alphabet | Fusional | Char 4-grams |
| Italian | ita | IE:Romance | Alphabet | Fusional | Word 1-grams |
| Portuguese | por | IE:Romance | Alphabet | Fusional | Word 1-grams |
| Romanian | ron | IE:Romance | Alphabet | Fusional | Char 4-grams |
| Spanish | spa | IE:Romance | Alphabet | Fusional | Word 1-grams |
| Japanese | jpn | Isolate | Logographic | Agglutinative | Char 3-grams |
| Korean | kor | Isolate | Logographic | Agglutinative | Char 4-grams |
| Arabic | ara | Semitic | Abjad | Root-and-Pattern | Word 1-grams |
| Hebrew | heb | Semitic | Abjad | Root-and-Pattern | Char 4-grams |
| Chinese | zho | Sino-Tibetan | Logographic | Analytic | Char 3-grams |
| Thai | tha | Tai-Kadai | Abugida | Analytic | Char 4-grams |
| Turkish | tur | Turkic | Alphabet | Agglutinative | Char 4-grams |
| Estonian | est | Uralic | Alphabet | Fusional | Word 1-grams |
| Finnish | fin | Uralic | Alphabet | Agglutinative | Char 4-grams |

| Name | Code | Family | Script | Morphology | Feature Type |
|---|---|---|---|---|---|
| Hungarian | hun | Uralic | Alphabet | Agglutinative | Char 4-grams |

There are some low-level differences between these corpora that we are not concerned with: the use of uppercase letters or punctuation or emojis, for example. The corpora are therefore preprocessed to normalize case, punctuation, numbers, email addresses, URLs, and other non-linguistic material that would make it easier to superficially distinguish between registers in this setting. Note that the corpora do not contain information about sentence segmentation, which means we are unable to evaluate sentence-based questions (i.e., Xu & He, 2018).

Several languages in this dataset have unique word segmentation patterns: Chinese (zho), Japanese (jpn), Thai (tha), and Tamil (tam). One side-effect of these word segmentation patterns, for example, is that the Tamil corpus from social media has different word boundaries than the Tamil corpus from Wikipedia. Rather than include language-specific tools in the pipeline, we instead remove all spaces from these corpora, in essence normalizing across styles of word segmentation. We then use only character-based features for these languages.

In order to maintain validity across many experiments, these corpora are divided into three subsets: training data, testing data, and validation data. We use the training data for feature selection and to determine the threshold for separating registers. Each of the parameters in the experiment is then evaluated on the testing data in order to determine, for each language, the best measure of corpus similarity. We then take these best measures and evaluate them on the held-out validation data. Thus, each language is evaluated on the validation corpus only once. This prevents us from over-fitting specific properties of that corpus. An additional experiment, described in Section 4, uses an independent out-of-domain corpus for feature selection. This allows us to measure the degree to which the corpus similarity measures depend on the specific registers being used in these experiments: would the measures continue to work if we had trained on Bible translations or movie subtitles? The basic finding is that using out-of-domain training data produces only a small decline in overall performance.

For each language, we draw samples from the three registers (TW, CC, WK). To evaluate the similarity measures we create 100 pairs of sub-corpora for each of the six possible combinations: TW-TW, TW-CC, TW-WK, CC-CC, CC-WK, WK-WK. There are no repeating pairs. This provides three populations of same-register pairs and three populations of different-register pairs, for a total of 600 pairs of sub-corpora per condition. As described below in more detail, the notion of accuracy in this setting involves using the similarity measures to predict, for each of these 600 pairs, whether the two corpora belong to the same or different registers.

A frequency-based approach uses bag-of-words features. Given a corpus, we create a frequency vector of length *n*, where *n* is the number of features being considered. We evaluate both word-based features (which use whitespace to divide units) and character-based features (which do not depend on whitespace in the same way). For each language for each parameter being evaluated we use a single feature space. This type of feature has also been used, for example, to identify distinct languages as well as distinct corpora from the same language (Seifart & Mundry, 2015). If we are evaluating Spanish with 5k word-based features, we use the same vocabulary of 5k words for each pair of corpora. Because the focus is on frequency differences within a language, this work is complementary to investigations of cross-linguistic word frequency distributions (Bentz, et al., 2017). Taken individually, these n-gram features represent a specific linguistic property; taken in the aggregate, they represent a potentially wide range of lexical, morphological, and syntactic properties. The focus is on evaluating performance in the aggregate rather than attempting to interpret the contribution of individual features.

The selection of features is based on frequency alone: we take the most common words in the training corpus. This means, for example, that the measures are not capturing information about lexical richness (i.e., Kubát & Milička, 2013; Shi & Lei, 2020) which focus on less common vocabulary features. From a different perspective, however, these high frequency features can provide an indicator for differences in syntactic patterns (i.e., Wan, et al., 2019). For the out-of-domain experiment discussed in Section 4.4, we test whether performing feature selection on an independent corpus from a different register reduces performance. A summary of the experimental conditions is given in Table 2.

**Table 2. Experimental Conditions**

| Corpus Size | N. Features | Feature Type | Measure |
|---|---|---|---|
| 10k words | 5k | Word 1-gram | Spearman $\rho$ |
| 30k words | 10k | Word 2-gram | $\chi^2$ |
| 50k words | 15k | Word 3-gram | Euclidean Distance |
| 100k words | 20k | Character 2-gram | Cosine Distance |
| 500k words | 25k | Character 3-gram | - |
| - | - | Character 4-gram | - |

To summarize, the basic experimental paradigm is to create a large number of unique pairs of sub-corpora for each language, where some pairs come from the same register and some from different registers. This allows us to evaluate the measures across different register boundaries.

We further divide the corpora into training data, testing data, and validation data to ensure that the large number of comparisons does not lead us to inflate the performance of the measures. A final experiment, using data from independent registers, provides an evaluation of out-of-domain feature selection in order to offer a further guarantee of robustness. This experimental design asks how robust corpus similarity measures remain across languages and across new corpora.

**4. Analysis**

This section presents the performance of corpus similarity measures by language using a held-out test set to provide a rigorous evaluation. We start by defining the notion of accuracy for a continuous measure (4.1). We then evaluate the relationship between corpus size and feature type (4.2) and corpus size and measure (4.3). We finish the section by asking whether the chosen parameters are specific to these registers by replicating the experiments using out-of-domain training data (4.4). The overall goal of these experiments is to determine whether corpus similarity as it has been defined in the literature extends beyond Indo-European languages.

*4.1. Threshold Values for Calculating Accuracy*

A corpus similarity measure provides a continuous representation of the difference between two input corpora. In order to calculate accuracy from this measure, we develop a threshold. Given two corpora, A and B, a similarity threshold is used to predict whether both A and B belong to the same register: values above the threshold are positive and values below the threshold are negative. We then report accuracy using this threshold on the held-out test corpus.

There are two variant algorithms for setting the threshold. The first, T1 (shown below), is the average similarity value across all types of pairs. This approach takes the global population of pairwise similarity, finds the mean value, and uses that mean value as the threshold: any pair above the average is predicted to be from the same register and any pair below the average is predicted to be from different registers. This version does not distinguish between same-register and different-register pairs when determining the threshold.

$$\text{avg}_v(\text{T1}) = \frac{1}{6} \sum_{i=1}^{6} \text{similarity value}_i$$

We contrast this with a variant that distinguishes between same-register and different-register pairs in the training data. This variant, T2, is shown below. We take the lowest average similarity for same-register pairs (for example, maybe CC-CC is the least homogenous register). Then we take the highest average similarity for different-register pairs (for example, maybe CC-WK are the most similar registers). The threshold is set halfway between these values.

$$\text{avg\_v(T2)} = \frac{1}{2}(\min(\text{similarity value}_{CC\_CC}, \text{similarity value}_{TW\_TW}, \text{similarity value}_{WK\_WK}) \\ + \max(\text{similarity value}_{CC\_TW}, \text{similarity value}_{CC\_WK}, \text{similarity value}_{TW\_WK}))$$

Which of these methods for setting the threshold values works best and do they actually capture the distinction between same-register and different-register similarity values? We survey a few languages in Figure 1 and Figure 2 to visualize the effect of the thresholds. In these figures, each column is a different language, with the table below the figure showing both the thresholds and their accuracy. The y-axis shows the computed similarity value, ranging from 0 (no similarity) to 1 (high similarity). Each dot represents a pair of two corpora. Blue dots represent corpora from the same register and orange dots represent corpora from different registers. These figures are based on Spearman's $\rho$ with 5k features per language; later sections justify this choice.

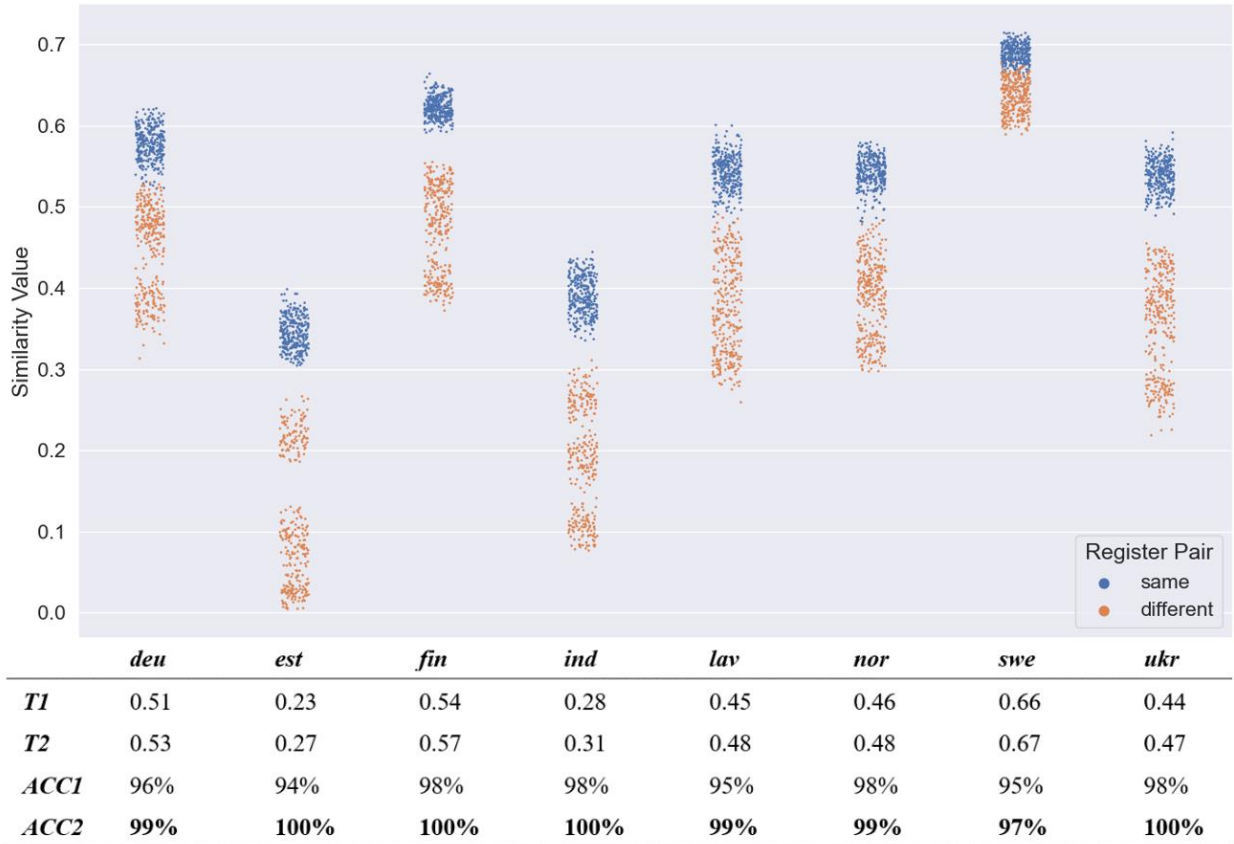

**Figure 1. Selected languages with different thresholds that have better accuracy using T2**

We see from Figure 1, first, that the same-register pairs always have a higher similarity (i.e., the blue dots are always above the orange dots). Thus, the corpus similarity measure is arranging the pairs of corpora as we would expect: data from the same register is the most similar. However, the gap between the two conditions varies by language. For example, in Estonian (est) and Finnish (fin), there is a clear separation between conditions. But in Swedish (swe) the two conditions meet, with some samples not clearly in one condition or another. This lack of separation is reflected in the accuracy measure, shown in the table below the figure: Estonian reaches 100% accuracy because the different-register pairs are always separated; but Swedish reaches only 97% accuracy because a few different-register pairs cross the threshold.

The second thing we notice from Figure 1 is that there is variation in the distribution of similarity values by language, an issue to which we return in Section 5. For example, Swedish (swe) has a dense cluster of pairs of corpora, all with values between 0.6 and 0.7. But Estonian (est) shows a much broader range of values, all with a lower tendency, from 0.0 to 0.4. This means that we cannot compare absolute similarity values across languages: 0.6 would be a very high similarity for Estonian but a very low similarity for Swedish. This is not a problem for the measures, however, because all comparisons take place within the same language. For the languages in Figure 1, the threshold selection in T2 performs better.

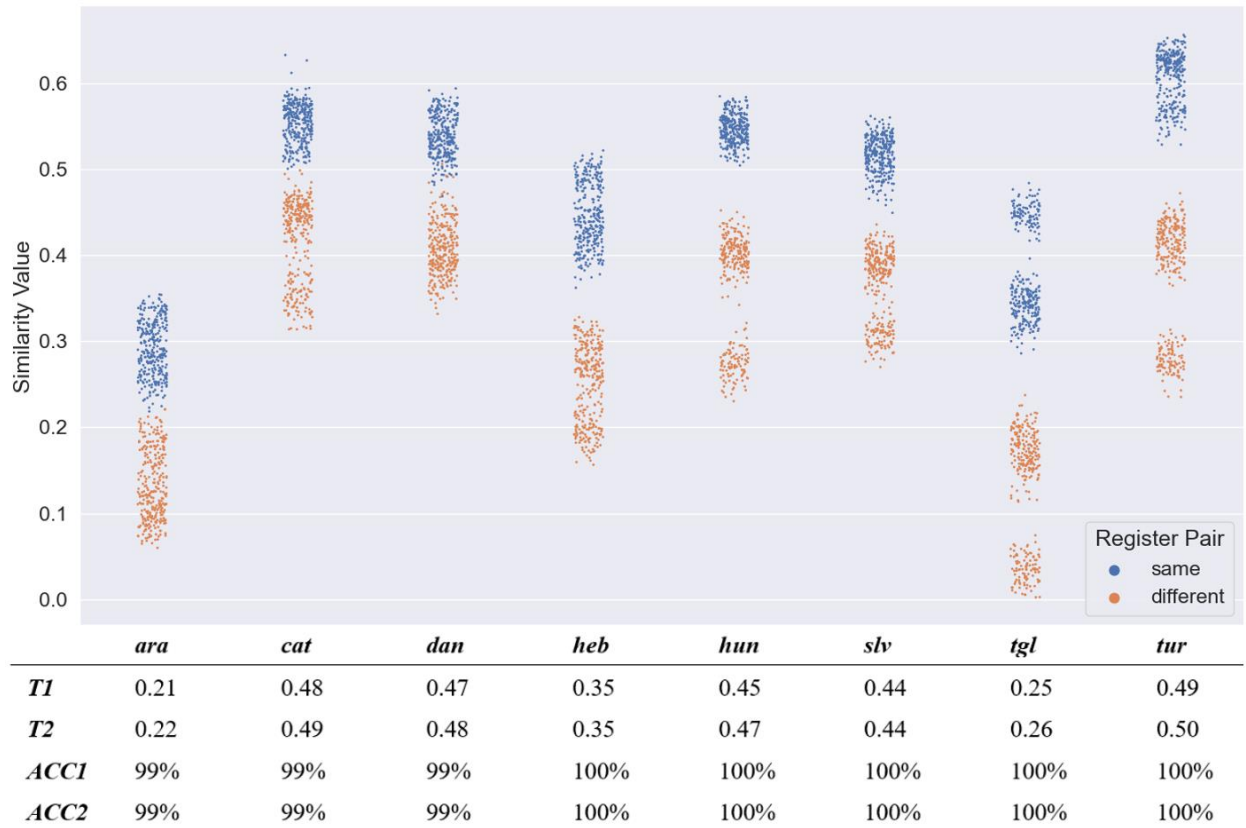

| | ara | cat | dan | heb | hun | slv | tgl | tur |
|---|---|---|---|---|---|---|---|---|
| *T1* | 0.21 | 0.48 | 0.47 | 0.35 | 0.45 | 0.44 | 0.25 | 0.49 |
| *T2* | 0.22 | 0.49 | 0.48 | 0.35 | 0.47 | 0.44 | 0.26 | 0.50 |
| *ACC1* | 99% | 99% | 99% | 100% | 100% | 100% | 100% | 100% |
| *ACC2* | 99% | 99% | 99% | 100% | 100% | 100% | 100% | 100% |

**Figure 2. Select languages with similar thresholds that show the same accuracy for both.**

There are other cases, however, in which the two methods perform equally well, as shown in Figure 2. We notice that in this selection of languages, there is a much clearer gap between same-register and different-register similarity values. Languages like Tagalog (tgl) and Turkish (tur) show a wide separation and other languages like Arabic (ara) have an intermediate area that is sparsely populated, meaning that few pairs fall into the center range. In these cases, the choice of a threshold value is much easier and both measures reach the same accuracy.

The point of this section is to visualize the distribution of similarity values, with a focus on the distinction between same-register pairs and different-register pairs. A corpus similarity measure is continuous. But we convert this continuous measure into a binary prediction using a simple threshold to determine when two corpora represent different registers.

*4.2. Corpus Size and Feature Type*

Having established a method for measuring the accuracy of different corpus similarity measures, we use this accuracy in predicting same-register vs different-register for validation purposes. The first question is to establish the relationship between corpus size (how much data we need for each sample) and the number of features (how many word or character sequences we include in the frequency vector). This relationship is shown in Figure 3 for a selection of four languages: Arabic (ara), German (deu), French (fra), and Farsi (fas). This selection includes different language families and types of morphology.

The y-axis represents corpus size, the number of words for each sample being compared (all comparisons are made with equal-sized samples). These range from 10k to 500k words: 10k (S1), 30k (S2), 50k (S3), 100k (S4), 500k (S5). Thus, the top row represents the smallest sample size and the bottom row the largest. The x-axis represents the type of feature: character 2-grams (C2) through word 3-grams (W3).

We notice, first, that each language has at least one type of feature which achieves high accuracy at a sample size of 10k words. Arabic has three such features, but other languages have only one. Given these results, we see that corpus similarity measures remain robust at relatively small sample sizes. In other words, larger samples are not necessary to achieve high accuracy. For most features, accuracy increases with more data per sample; this is not true, however, for character bigrams in Arabic.

Given the full experimental results (not shown), we divide languages into four groups based on the best type of feature. This was shown in Table 1. If a language (such as Arabic) achieves high accuracy across more than one type of feature, we choose the feature type which is most common across other languages. Most languages prefer character 4-grams (27). A number of other languages prefer word 1-grams (9). Only a few remaining languages prefer character 3-grams (2) or word 2-grams (1). For the remaining experiments, we fix the type of feature for each language in accordance with the values shown in Table 1 (the best performing features). And we report results with corpus size limited to 25k words. The reason for limiting these hyper-parameters is that we see high accuracy in these particular configurations.

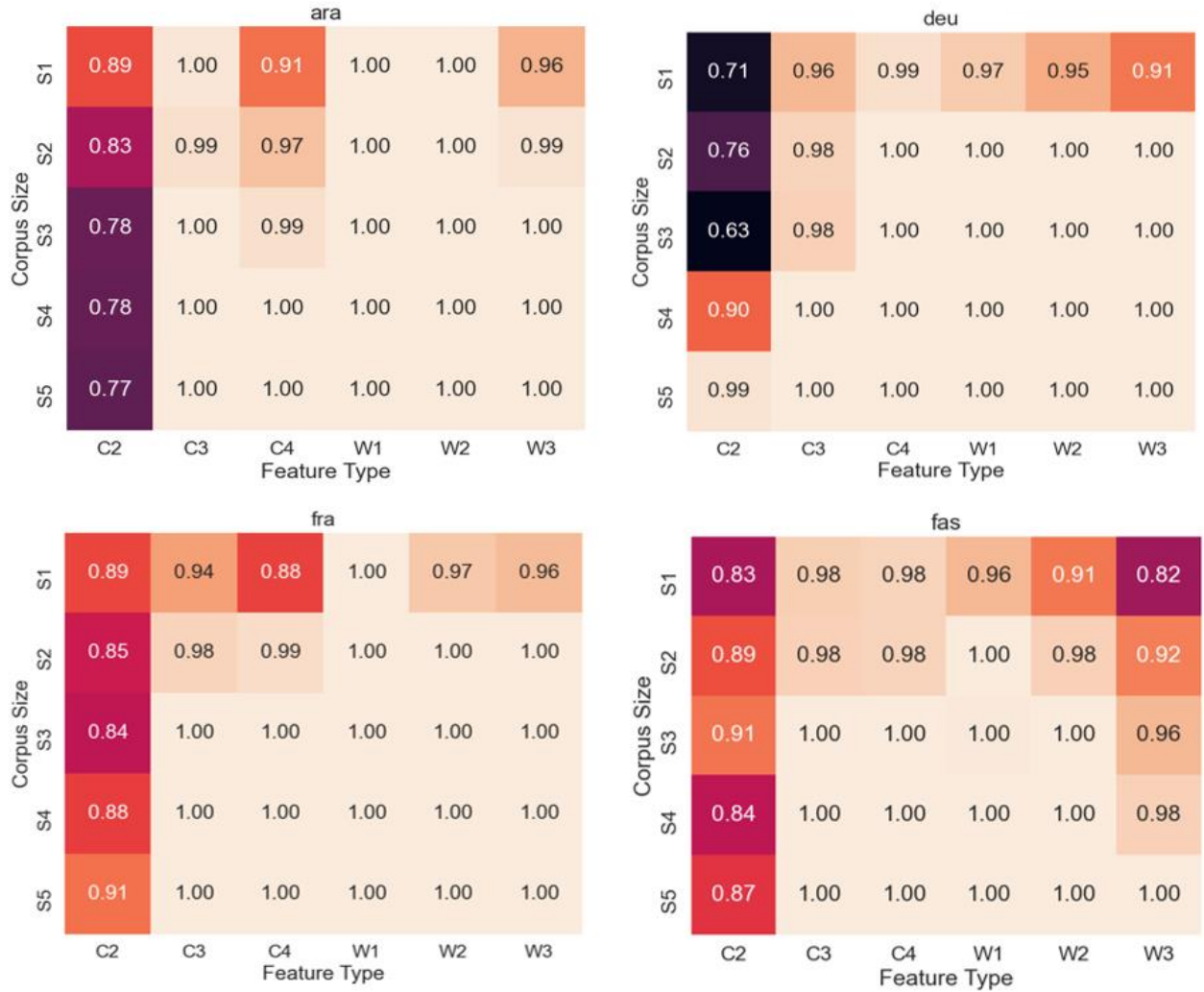

**Figure 3. Accuracy results based on corpus size vs feature type show that every language has its own configuration for producing the best performance.**

*4.3 Comparing Measures of Similarity*

We turn now to a comparison of similarity measures, with the feature type fixed by language given the previous experiments. This is shown in Table 4 with samples containing 25k words. The first thing we notice is that, as per previous work, only Spearman's $\rho$ and $\chi^2$ perform consistently well. For example, Euclidean distance performs poorly on Hindi (hin) and Japanese (jpn). And Cosine distance performs poorly on Estonian (est) and Polish (pol). Even though these two measures perform well on many languages, they do not outperform the best measures in previous work. Therefore, we exclude them from further consideration.

Should we prefer Spearman's $\rho$ or $\chi^2$? Both measures achieve at least 99% accuracy on 28 different languages (but not the same 28). There are three languages where Spearman is significantly worse (Bulgarian, Swedish, Tagalog); and five languages where Spearman is significantly better (Farsi, Hindi, Japanese, Spanish, Urdu). There is only one language (Tagalog) which would fall below 95% accuracy if we used Spearman's $\rho$. As a result of these findings, we use Spearman's $\rho$ across all 39 languages. The results for smaller samples (10k words) are provided in the supplementary material.

Why not mix-and-match similarity measures in the same way that we mix different types of features? There are two reasons to prefer Spearman's $\rho$: First, it is not dependent on a fixed sample size. On the other hand, $\chi^2$ is dependent on the length of each sample and thus cannot be used under conditions with uncertain corpus lengths (Kilgarriff, 2001). The second advantage of Spearman's $\rho$ is that it always falls between 0 and 1, making it possible to visualize thresholds across languages. Because $\chi^2$ produces output values with no maximum, the comparison of different languages or different registers is quite difficult. For example, if we used $\chi^2$ only for a few languages like Tagalog, it would not be possible to compare those similarity measures to the measures for other languages. For these reasons, we use Spearman's $\rho$ across all languages.

**Table 4. Accuracy by Similarity Measure with 25k word samples**

|     | Spearman | $\chi^2$ | Cosine | Euclidean |     | Spearman | $\chi^2$ | Cosine | Euclidean |
| --- | --- | --- | --- | --- | --- | --- | --- | --- | --- |
| **ara** | 100% | 100% | 100% | 99.8% | **lav** | 100% | 100% | 100% | 100% |
| **bul** | 99.7% | 100% | 100% | 100% | **nld** | 100% | 100% | 100% | 100% |
| **cat** | 100% | 100% | 99.3% | 98.4% | **nor** | 100% | 100% | 100% | 100% |
| **ces** | 100% | 100% | 99.5% | 100% | **pol** | 100% | 100% | 100% | 100% |
| **dan** | 100% | 100% | 100% | 100% | **por** | 100% | 100% | 98.7% | 100% |
| **deu** | 100% | 100% | 100% | 100% | **ron** | 100% | 100% | 100% | 100% |
| **ell** | 100% | 100% | 99.8% | 100% | **rus** | 100% | 100% | 100% | 100% |
| **eng** | 100% | 98.7% | 97.7% | 97.7% | **slv** | 100% | 100% | 80.0% | 62.4% |
| **est** | 100% | 100% | 71.7% | 100% | **spa** | 100% | 100% | 100% | 100% |

|     | Spearman | $\chi^2$ | Cosine | Euclidean |     | Spearman | $\chi^2$ | Cosine | Euclidean |
| --- | --- | --- | --- | --- | --- | --- | --- | --- | --- |
| **fas** | 99.5% | 100% | 98.4% | 97.9% | **swe** | 100% | 100% | 100% | 100% |
| **fin** | 100% | 100% | 100% | 100% | **tam** | 98.0% | 100% | 98.2% | 100% |
| **fra** | 100% | 100% | 100% | 100% | **tel** | 100% | 100% | 100% | 100% |
| **glg** | 100% | 99.8% | 95.6% | 89.4% | **tgl** | 90.2% | 100% | 100% | 100% |
| **heb** | 100% | 100% | 100% | 93.5% | **tha** | 99.5% | 97.7% | 97.9% | 97.7% |
| **hin** | 97.1% | 98.9% | 86.0% | 95.9% | **tur** | 100% | 100% | 98.9% | 99.8% |
| **hun** | 100% | 100% | 100% | 100% | **ukr** | 100% | 100% | 100% | 100% |
| **ind** | 100% | 100% | 100% | 100% | **urd** | 100% | 100% | 99.5% | 100% |
| **ita** | 100% | 100% | 100% | 100% | **vie** | 100% | 100% | 100% | 100% |
| **jpn** | 100% | 100% | 98.2% | 97.7% | **zho** | 100% | 100% | 96.4% | 94.6% |
| **kor** | 100% | 100% | 100% | 100% |     |     |     |     |     |

## 4.4. In-Domain vs Out-of-Domain Evaluation on the Validation Corpus

The previous experiments have learned both features and thresholds from the training corpora and evaluated these settings on the testing corpora. We have now determined the best parameters: Spearman's $\rho$ with 5k features, with feature type varying by language. The feature space and threshold for distinguishing between same-register and different-register pairs are fixed using the training data. In this section we evaluate these hypothesized best measures on the validation corpora, data which has remained unseen until now. This final step ensures the robustness of our results.

A further question that we might ask is whether these corpus similarity measures depend on the three registers that we have used for training (social media, web macro-register, non-fiction encyclopedia articles). If we trained the model with Bible translations or movie subtitles, would the measures be comparable? The reason this question matters is because we have used the same feature space for each language. In other words, we find the 5k most frequent features (words or character sequences) in the training data. Those features form the feature vector for each sample, even if some features are missing from both samples. There is a possibility that, as we move away from the registers represented in the training data, the number of missing features increases so that these measures would fail on new domains.

To evaluate this possibility, we measure the accuracy for each language in two conditions: *First*, using the previous in-domain training data for feature selection; *Second*, using an independent corpus from different registers for feature selection. These independent corpora contain a random mix of Bible translations, news commentary articles, and movie subtitles (Tiedemann, 2012; Christodoulopoulos & Steedman, 2015). If the measures are over-fitting the training/testing data, then the in-domain performance on the validation data will be significantly lower than seen in Table 4. If the measures are applicable only to the registers represented in the training data, the out-of-domain performance will be significantly lower. The results, shown in Table 5, provide a final evaluation of corpus similarity measures across languages. The *Difference* column shows the change between in-domain and out-of-domain feature selection. Changes less than 1% are removed; decreased accuracies of 5% or more are indicated in bold.

No language falls below 95% accuracy for either in-domain or out-of-domain feature selection (with samples of 25k words). This provides a robust baseline for these measures. In other words, given corpora of at least 25k words we can have high confidence in these measures. Overall, these results show that the best corpus similarity measures are robust across all languages, even if a different set of registers is involved. This indicates that the measures are not dependent on specific registers but can be taken as generalized measures of similarity.

**Table 5. In-Domain and Out-of-Domain Performance on the Validation Set, 25k Word Samples**

| Language | In Domain | Out Domain | Difference | Language | In Domain | Out Domain | Difference |
|---|---|---|---|---|---|---|---|
| ara | 100% | 100% | | lav | 100% | 100% | |
| bul | 99.2% | 99.8% | | nld | 100% | 100% | |
| cat | 100% | 100% | | nor | 100% | 100% | |
| ces | 100% | 100% | | pol | 100% | 100% | |
| dan | 100% | 100% | | por | 100% | 99.7% | |
| deu | 100% | 100% | | ron | 100% | 100% | |
| ell | 100% | 100% | | rus | 100% | 100% | |
| eng | 100% | 99.8% | | slv | 100% | 100% | |
| est | 100% | 100% | | spa | 100% | 99.8% | |

| Language | In Domain | Out Domain | Difference | Language | In Domain | Out Domain | Difference |
|---|---|---|---|---|---|---|---|
| fas | 99.8% | 100% | | swe | 100% | 96.1% | -3.90% |
| fin | 100% | 100% | | tam | 100% | 100% | |
| fra | 99.8% | 100% | | tel | 100% | 99.0% | -1.00% |
| glg | 100% | 100% | | tgl | 97.4% | 100% | +2.60% |
| heb | 100% | 100% | | tha | 97.4% | 99.5% | +2.10% |
| hin | 97.9% | 100% | +2.10% | tur | 100% | 100% | |
| hun | 100% | 100% | | ukr | 100% | 100% | |
| ind | 100% | 100% | | urd | 100% | 99.5% | |
| ita | 99.8% | 96.3% | -3.50% | vie | 98.7% | 99.8% | +1.10% |
| jpn | 100% | 100% | | zho | 100% | 100% | |
| kor | 100% | 96.3% | -3.70% | | | | |

## 5. Discussion

The previous experiments have shown that corpus similarity measures across 39 languages provide highly accurate predictions on held-out validation data with both in-domain and out-of-domain feature selection. This shows both (i) that corpus similarity measures generalize across languages and (ii) that their accuracy is not specific to each data set. In this section we take a closer look at three important questions: First, why do some languages have lower performance with out-of-domain feature selection? Second, how much variation do we observe both within and between specific registers across languages? The question is important for understanding what happens when we encounter a macro-register like web pages. Third, we experiment with low-resource Austronesian languages to determine whether the behaviour of select languages is representative of the larger family they belong to. Fourth, we end by examining a selection of non-digital registers in English to ensure that these results also extend to more traditional corpora.

*5.1. Overlapping Features Between In-Domain and Out-of-Domain Corpora*

Why do some languages have reduced performance given out-of-domain feature selection? For example, Korean (kor) sees better accuracy with in-domain features than out-of-domain features (100% vs 96.3%). One possibility is that there is a low overlap between the features, with the out-of-domain features not being present for the in-domain validation data. In this case, only 48% of the feature set is shared between in-domain and out-of-domain feature selection (c.f., the full table of feature overlap in the supplementary material). In other words, if we choose the 5k most frequent features on the out-of-domain corpus, roughly half of those features are not in the top 5k features derived from the in-domain corpus. This is contrasted with Chinese (zho), which has a similarly low feature overlap (46.5%), but does not show a drop in accuracy (both have 100% accuracy). If low overlap were the primary cause of low out-of-domain accuracy, then most low overlap languages would have reduced accuracy. Of the 17 languages with less than 75% feature overlap, only two have below 99% out-of-domain accuracy (Korean and Italian). This implies that feature overlap is not the main factor for reduced out-of-domain accuracy. These two languages also are from different families, with different writing systems and types of morphology.

Returning to the idea of feature overlap, there is a second question beyond the impact on accuracy: whether or not there is a reduction in performance, why do some languages have low feature overlap? Here we focus on the 17 languages with below 75% overlap. First, we notice that all three logographic script languages are below 50% overlap (Chinese, Japanese, Korean). Given the increased inventory of symbols, this low overlap is not surprising. In fact, all five languages which use spaceless preprocessing (including also Tamil and Thai) have lower overlap. This means that this method for normalizing differences in word segmentation reduces feature overlap, although it does not reduce accuracy. We also notice that all abugida languages (where consonant-vowel sequences are written together) fall below the 75% overlap threshold. This indicates that the main driver of low feature overlap is the type of script, rather than, for example, the type of morphology. At a sample size of 25k words, no language falls below 95% accuracy with out-of-domain features, however, which indicates that lower feature overlap does not cause lower prediction accuracy for corpus similarity measures.

*5.2. Similarity Distribution Across Registers*

The previous experiments have focused on a distinction between same-register and different-register pairs, in effect averaging across three distinct registers (social media, web, and non-fiction encyclopedia articles). In this section, we take a closer look at the distribution of similarity values across registers. This allows us to find further explanations for the accuracy results in Table 5. As noted above, the web is actually a macro-register with a number of potentially distinct sub-registers. Do these sub-registers create increased heterogeneity that is discovered by the similarity measures? A more specific question is whether we see an impact from the internal variation caused by web pages representing a macro-register.

To analyze the distribution of similarity values, we start by extracting 600 pairs of corpora from the validation set for each language; as before, this includes 100 pairs for each condition (c.f., Table 3). The similarity measures have the same hyper-parameters used for Table 5: 5k features of the specified type, either in-domain or out-of-domain, compared using Spearman's $\rho$. Thus, these experiments correspond to the results in Table 5. We use violin plots to show the distribution for each language in Figures 5 through 7. These figures focus on three representative languages: Italian (ita), Hindi (hin), and English (eng). For each language the figures contain two facets: feature type (in-domain and out-of-domain). Within each facet, the distribution of same-register pairs is shown on the left (TW, CC, WK) and the distribution of different-register pairs is shown on the right (TW-CC, WK-CC, TW-WK). For each violin plot the y-axis is the similarity value, with higher pairs being more similar and lower pairs being less similar. A wider violin plot shows that more pairs are found at that point of similarity, with the mean represented by a white dot in the center of the plot. A horizontal line is added which represents the threshold selected using the T2 method described above.

We notice, first, that the same-register pairs (on the left) have higher similarity than the different-register pairs (on the right). This is expected, of course, given the high accuracy for the measures, which requires a fixed threshold between the two classes. But we also notice that there is variation in the internal similarity (homogeneity) for same-register pairs. Social media (TW) is the most homogenous across each language because its average self-similarity is higher.

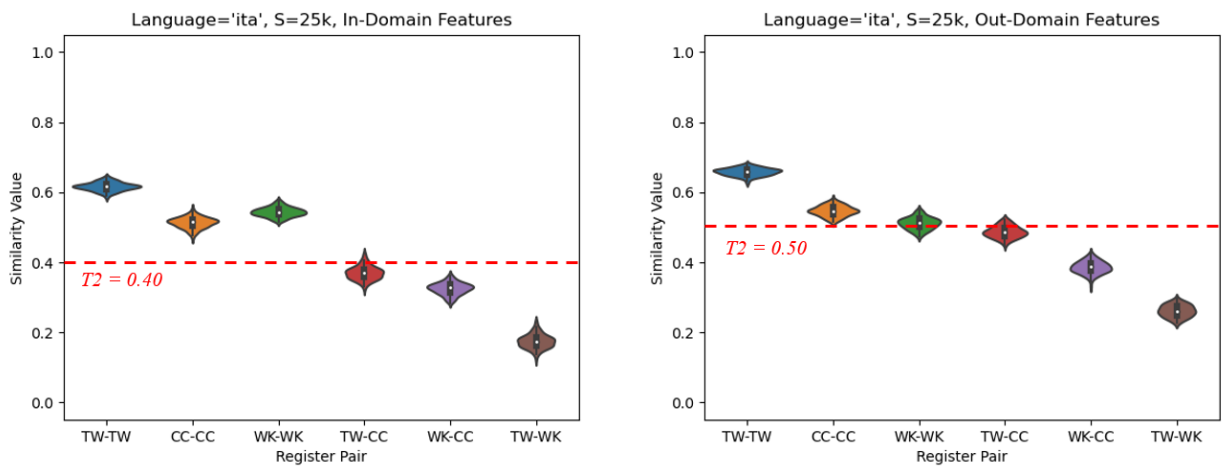

**Figure 5. Distribution of Similarity Values, Italian**

We also notice that, for each language, the variation between social media (TW) and Wikipedia (WK) is the greatest. In other words, these two registers are the least similar across languages. For Hindi (hin), unlike Italian and English, the gap between the web and social media is quite similar to the gap between Wikipedia and social media. Thus, the relationships between each pair of registers may not be constant across languages in the same way that the ordering is relatively constant.

If we look more closely at Italian, in Figure 5, we see that the difference in accuracy between in-domain and out-of-domain feature selection is driven by the same-register similarity of Wikipedia. With in-domain features, Wikipedia is clearly distinct from the different-register pairs on the right side of the figure. But, with out-of-domain features, the internal homogeneity of Wikipedia is much lower. This indicates that those features which do not overlap are specific to the Wikipedia register.

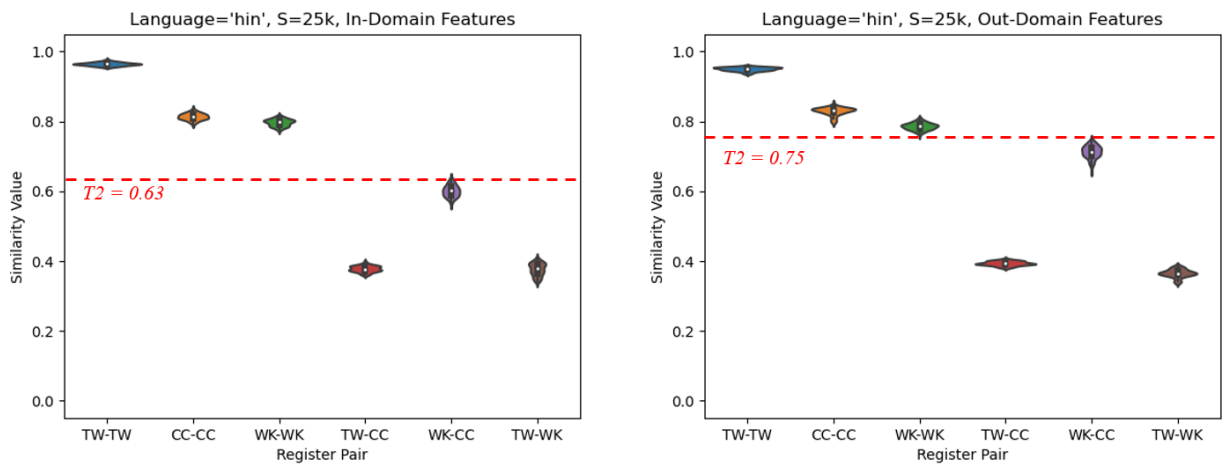

**Figure 6. Distribution of Similarity Values, Hindi**

These violin plots also show the variation in each pair of corpora. For example, looking at the English graph in Figure 7, we notice that the values for social media (TW) are both very similar (high) but also very centered around the mean (the violin plot is wide). On the other hand, the values comparing social media and Wikipedia are both quite dissimilar (low) but also stretched across a range of values. This variation in each distribution is another indicator of heterogeneity within a particular register.

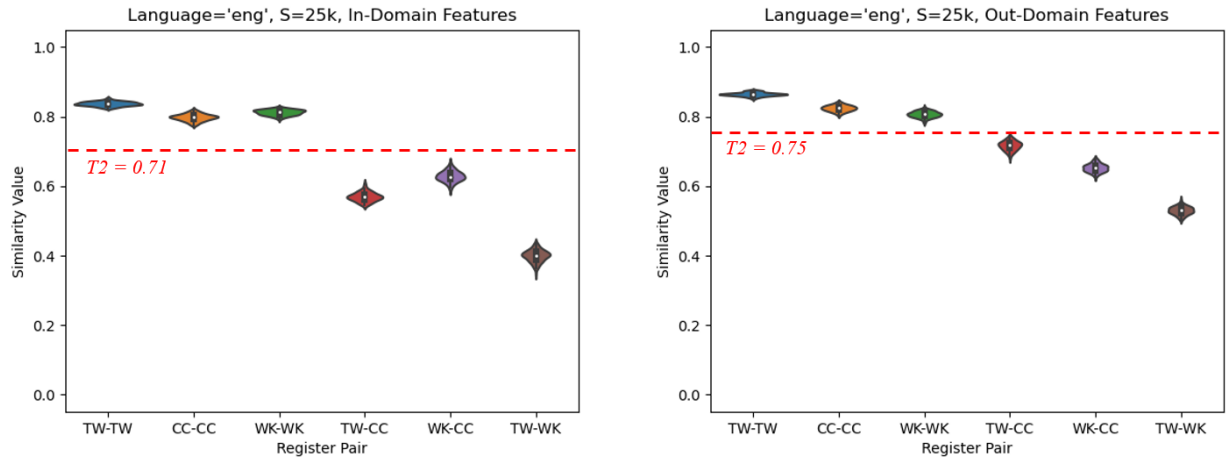

**Figure 7. Distribution of Similarity Values, English**

The accuracy evaluation focuses on finding a single threshold (i.e., a horizontal line here) that separates same-register and different-register pairs. The practical value of corpus similarity measures, however, is in the full range of values. We notice, for example in reference to the Italian data in Figure 5, that the accuracy metric is lowered by cases like WK-WK and TW-CC, two sets of pairs that have comparable distributions of similarity measures.

This section has explored the distribution of similarity measures further, with a focus on register-by-register results. We have seen that, in addition to making accurate predictions, corpus similarity measures provide a great deal of information about register variation itself when viewed in their continuous form. A complete set of language-specific figures is provided in the supplementary material. We also see that the potential internal variation caused by sub-registers within the web corpora does not pose a problem for corpus similarity measures.

*5.3. Application to Less Common Languages*

This section examines a further question: how robust is this framework for corpus similarity when it is applied to previously unseen, less common languages? In this case, we look at eight Austronesian languages, half of which belong to the Polynesian sub-family. Two of these languages are relatively high-resource languages and have appeared in the previous experiments (Indonesian and Tagalog). But the rest are relatively low-resource languages that have not been included in the previous experiments. The question here is whether the performance on the high-resource members of this family actually extends to the low-resource members as well.

The low-resource languages here do not have the same three comparable corpora used above (TW, CC, WK). Instead, there is an idiosyncratic mix of registers for each language. We evaluate corpus similarity in these language in a more difficult setting, with 10k word samples using character 4-grams. Because there is less data per language, we take a cross-validation approach and report the accuracies for each fold in Table 6. The number of registers per language is also shown, ranging from a wider prediction task (five registers for Tagalog) to a narrower prediction task (two registers for Malagasy). The performance is quite high, so that similarity measures are still able to predict register differences even in these low-resource languages.

*Table 6. Cross-Validation Accuracy for Less Common Austronesian Languages*

| Name | Family | Registers | N. Test | CV 1 | CV 2 | CV 3 | CV 4 | CV 5 |
|---|---|---|---|---|---|---|---|---|
| Cebuano | Austronesian | 3 | 331 | 100% | 100% | 100% | 100% | 100% |
| Malagasy | Austronesian | 2 | 243 | 100% | 100% | 100% | 100% | 100% |
| Indonesian | Austronesian | 4 | 310 | 100% | 100% | 100% | 100% | 100% |
| Tagalog | Austronesian | 5 | 328 | 100% | 100% | 100% | 100% | 100% |
| Hawaiian | Polynesian | 2 | 31 | 100% | 100% | 100% | 100% | 100% |
| Samoan | Polynesian | 3 | 199 | 100% | 100% | 100% | 100% | 100% |
| Te Reo Māori | Polynesian | 2 | 31 | 100% | 100% | 100% | 100% | 100% |
| Tongan | Polynesian | 3 | 185 | 100% | 100% | 100% | 100% | 100% |

This experiment is important because it shows that the performance within a language family extends from the high-resource languages previously considered to other low-resource languages in that family. The experiment also shows that the previous results do not depend on the specific three registers being tested. Here a different set of registers is used, still showing a high level of accuracy. While the previous comparable corpora were important for establishing cross-linguistic patterns, these results show that those patterns are not dependent on these specific corpora.

*5.4. Application to Non-Digital Registers*

In this section we conduct an additional experiment to determine whether these results generalize beyond the digital registers so far considered. In other words, it is possible although unlikely that these results are dependent on the fact that all three registers considered so far are drawn from digital sources. To test this, we bring together six corpora from written non-digital registers: the European Central Bank Corpus (*Bank:* Tiedemann, 2012); translations of the Bible (*Bible:* Christodoulopoulos & Steedman, 2015); a corpus of translated books (*Books:* Tiedemann, 2012); the GlobalVoices corpus of news and commentary articles (*News:* Tiedemann, 2012); the EuroParl corpus of parliamentary proceedings (*EuroParl:* Tiedemann, 2012); and a corpus of TedTalks (*Ted:* Reimers & Gurevych, 2020). This experiment thus represents a wide variety of non-digital registers in English.

We compare the similarity for each combination of these six registers using the above settings for English: 5k character four-grams (the same inventory) compared using Spearman's $\rho$. The sample size is 10k words per observation. We create 100 random pairs of sub-corpora for each pair of registers, the same experimental design used above. The threshold for predicting that two pairs come from different registers is calculated once, using the T2 equation described above. We evaluate this threshold in two conditions: one-vs-one (i.e., distinguishing samples from *Parliament* from samples from *Bank*) and one-vs-all (i.e., distinguishing samples from *Parliament* from all other registers). This experiment thus allows us to evaluate how well these results generalize beyond digital registers as well as to a larger inventory of registers.

*Table 7. Corpus Similarity in English Beyond Digital Registers, Accuracy*

|  | *Bank* | *Bible* | *Books* | *News* | *EuroParl* | *Ted* | 1-vs-ALL (700 pairs) |
|---|---|---|---|---|---|---|---|
| *Bank* |  | 100% | 100% | 100% | 100% | 100% | 100% |
| *Bible* |  |  | 100% | 100% | 100% | 100% | 100% |
| *Books* |  |  |  | 100% | 100% | 100% | 100% |
| *News* |  |  |  |  | 100% | 100% | 100% |
| *EuroParl* |  |  |  |  |  | 100% | 100% |
| *Ted* |  |  |  |  |  |  | 100% |

The results, shown in Table 7, are robustly high across registers. While some pairs of registers are more similar (for example, EuroParl and Bank are the most similar), it remains the case that a simple threshold and Spearman's $\rho$ are able to distinguish between a larger inventory of non-digital registers. These results thus show that the previous experiments generalize beyond digital contexts.

# 6. Conclusions

This paper has explored corpus similarity measures across 39 diverse languages. These measures have previously been studied almost exclusively in English; the languages in this paper, however, represent many language families, morphological systems, and writing systems. This is an important contribution because the robust performance across these experiments shows that corpus similarity measures are not dependent on *ad hoc* or non-linguistic features specific to individuals sets of data. In other words, they generalize within and across languages. This is further shown by the robust accuracy obtained even with out-of-domain features derived from independent registers for each language and when the measures are evaluated on previously unseen low-resource languages or on new sets of register-specific corpora.

The experiments in this paper show that high accuracy is possible on a task of predicting whether two samples come from the same register or from different registers. The basic conclusion, then, is that frequency-based corpus similarity measures remain robust across languages. This is true even with out-of-domain feature selection and is verified on a separate held-out validation corpus. While the discussion has focused on certain languages, the supplementary material contains the full results for further inspection.

Further, these measures are robust across languages when using a pre-defined feature space. We are thus able to release a Python package which contains everything necessary for using these measures, from preprocessing steps to the fixed n-gram features. This package is publicly available[1]. The important contribution of these experiments is to show that corpus similarity measures depend on properties of natural language data and do not depend on particular writing systems, particular types of morphology, or other properties of a single language family like Indo-European. In short, the experiments in this paper show that corpus similarity measures generalize well across new languages and across new corpora within a single language.

The number of corpora available for linguistic analysis has been rapidly increasing. At the same time, the size of individual corpora has also been growing, with many corpora now containing billions of words. Given this situation, how do we select which corpus to analyze? Does a syntactic analysis of news articles extend to Wikipedia articles? Does a lexical analysis of social media data extend to internet forums or text messages? By comparing corpora in the aggregate (Kilgarriff, 2001), corpus similarity measures allow us to systematically generalize corpus-based linguistic analysis. While previous work has been largely restricted to English, the experiments in this paper have shown that the same approach can be used on a diverse set of languages. This, in turn, is an important step for making corpus linguistics a multi-lingual discipline.

---

[1] https://github.com/jonathandunn/corpus_similarity